# Computing FIRST and FOLLOW Functions for Feature-Theoretic Grammars

Arturo Trujillo
Computer Laboratory
University of Cambridge
Cambridge CB2 3QG, England
iat@cl.cam.ac.uk

## Abstract

This paper describes an algorithm for the computation of FIRST and FOLLOW sets for use with feature-theoretic grammars, in which the value of the sets consists of pairs of feature-theoretic categories. The algorithm preserves as much information from the grammars as possible, using negative restriction to define equivalence classes. Addition of a simple data structure leads to an order of magnitude improvement in execution time over a naive implementation.

## 1 Introduction

The need for efficient parsing is a constant one in Natural Language Processing. With the advent of feature-theoretic grammars, many of the optimization techniques that were applicable to Context Free (CF) grammars have required modification. For instance, a number of algorithms used to extract parsing tables from CF grammars have involved discarding information which otherwise would have constrained the parsing process, Briscoe and Carroll (1993). This paper describes an extension to an algorithm that operates over CF grammar to make it applicable to feature-theoretic ones. One advantage of the extended algorithm is that it preserves as much of the information in the grammar as possible.

### 1.1 FIRST and FOLLOW

In order to make more efficient parsers, it is sometimes necessary to preprocess (compile) a grammar to extract from it top-down information to guide the search during analysis. The first step in the preprocessing stage of several compilation algorithms requires the solution of two functions normally called FIRST and FOLLOW. Intuitively, $FIRST(X)$ gives us the terminal symbols that may appear in initial position in substrings derived from category $X$. $FOLLOW(X)$ gives us the terminals which may immediately follow a substring of category $X$. For example, in the grammar S → NP VP; NP → det noun; VP → vtra NP, we get:

$FIRST(S) = FIRST(NP) = \{det\}$,
$FIRST(VP) = \{vtra\}$,
$FOLLOW(NP) = \{vtra, \$\}$,
$FOLLOW(S) = FOLLOW(VP) = \{\$\}$ (\$ marks end of input)

These two functions are important in a large range of algorithms used for constructing efficient parsers. For example the LR-parser construction algorithm given in Aho et al. (1986:232) uses FIRST to compute item closure values. Another example is the computation of the ∠∗ relation which is used in the construction of generalized left-corner parsers, Nederhof (1993); this relation is effectively an extension of the function FIRST.

# 2 Computing FIRST and FOLLOW

We propose an algorithm for the computation of FIRST values which handles feature-theoretic grammars without having to extract a CF backbone from them; the approach is easily adapted to compute FOLLOW values too. An improvement to the algorithm is presented towards the end of the paper. Before describing the algorithm, we give a well known procedure for computing FIRST for CF grammars (taken from Aho et al. (1986:189), where $\epsilon$ is the empty string):

"To compute $FIRST(X)$ for all grammar symbols $X$, apply the following rules until no more terminals or $\epsilon$ can be added to any FIRST set.

1. If X is terminal, then $FIRST(X)$ is $X$.

2. If $X \to \epsilon$ is a production, then add $\epsilon$ to $FIRST(X)$.

3. If $X$ is nonterminal and $X \to Y_1 Y_2 ... Y_k$ is a production, then place $a$ in $FIRST(X)$ if for some $i$, $a$ is in $FIRST(Y_i)$, and $\epsilon$ is in all of $FIRST(Y_1)$ ... $FIRST(Y_{i-1})$; that is, $Y_1...Y_{i-1} \stackrel{*}{\Rightarrow} \epsilon$. If $\epsilon$ is in $FIRST(Y_j)$ for all $j = 1, 2,..., k$, then add $\epsilon$ to $FIRST(X)$.

Now, we can compute FIRST for any string $X_1 X_2...X_n$ as follows. Add to $FIRST(X_1 X_2...X_n)$ all of the non-$\epsilon$ symbols of $FIRST(X_1)$. Also add the non-$\epsilon$ symbols of $FIRST(X_2)$ if $\epsilon$ is in $FIRST(X_1)$, the non-$\epsilon$ symbols of $FIRST(X_3)$ if $\epsilon$ is in both $FIRST(X_1)$ and $FIRST(X_2)$, and so on. Finally, add $\epsilon$ to $FIRST(X_1 X_2...X_n)$ if, for all $i$, $FIRST(X_i)$ contains $\epsilon$."

This algorithm will form the basis of our proposal.

# 3 Compiling Feature-Theoretic Grammars

## 3.1 Equivalence Classes

The main reason why the above algorithm cannot be used with feature-theoretic grammars is that in general the number of possible nonterminals allowed by the grammar is infinite. One of the simplest ways of showing this is where a grammar accumulates the orthographic representation of its terminals as one of its feature values. It is not difficult to see how one can have an infinite number of NPs in such a grammar:

NP[orth: the dog]
NP[orth: the fat dog]
NP[orth: the big fat dog], etc.

This means that $FIRST(NP[\text{orth: the dog}])$ would have a different value to $FIRST(NP[\text{orth: the fat dog}])$ even though they share the same leftmost terminal. That is, the feature structure for the substring "det adj noun" will be different to that for "det noun" even though they have the same starting symbol. This point is important since similar situations arise with the subcategorization frame of verbs and the semantic value of categories in contemporary theories of grammar, Pollard and Sag (1987). Without modification, the algorithm above would not terminate.

The solution to this problem is to define a finite number of equivalence classes into which the infinite number of nonterminals may be sorted. These classes may be established in a number of ways; the one we have adopted is that presented by Harrison and Ellison (1992) which builds on the work of Shieber (1985): it introduces the notion of a negative restrictor to define equivalence classes. In this solution a predefined portion of a category (a specific set of paths) is discarded when determining whether a category belongs to an equivalence class or not. For instance, in the above example we could define the negative restrictor to be {orth}. Applying this negative restrictor to each of the three NPs above would discard the information in the 'orth' feature to give us three equivalent nonterminals. It is clear that the restrictor must be such that it discards features which in one way or another give rise to an infinite number of nonterminals. Unfortunately, termination is not guaranteed for all restrictors, and furthermore, the best restrictor cannot be chosen automatically since it depends on the amount of grammatical information that is to be preserved. Thus, selection of an appropriate restrictor will depend on the particular grammar or system used.

## 3.2 Value Sharing

Another problem with the algorithm above is that reentrancies between a category and its FIRST and FOLLOW values are not preserved in the solution to these functions; this is because the algorithm assumes atomic symbols and these cannot encode explicitly shared information between categories. For example, consider the following naive grammar:

| S $\Rightarrow$ | NP[agr: X] VP[agr: X] |
| VP[agr: X] $\Rightarrow$ | Vint[agr: X] |
| NP[agr: X] $\Rightarrow$ | Det N[agr: X] |

We would like the solution of $FOLLOW(N)$ to include the binding of the 'agr' feature such that the value of FOLLOW resembled: $FOLLOW(N[agr:X]) = Vint[agr:X]$. But the algorithm above, even with a restrictor, would not preserve such a binding since the addition of a new category to $FOLLOW(N)$ is done independently of the bindings between the new category and $N$.

## 4 The Basic Algorithm

We propose an algorithm which, rather than construct a set of categories as the value of FIRST and FOLLOW, constructs a set of pairs each of which represents a category and its FIRST or FOLLOW category, with all the correct bindings explicitly encoded. For instance, for the above example, the pair (VP[agr: X], Vint[agr: X]) would be in the set representing the value of the function FIRST. In the next section the algorithm for computing FIRST is described; computation of FOLLOW proceeds in a similar fashion.

### 4.1 Solving FIRST

When modifying the algorithm of Section 2 we note that each occurrence of a category in the grammar is potentially distinct from every other category. In addition, for each category we need to remember all the reentrancies between it and the daughters within the rule in which it occurs. Finally, we assume that any category in a rule which can unify with a lexical category is marked in some way, say by using the feature-value pair 'ter: +', and that non-terminal categories must unify with the mother of some rule in the grammar; the latter condition is necessary because the algorithm only computes the solution of FIRST for lexical categories or for categories that occur as mothers.

In computing FIRST we iterate over all the rules in the grammar, treating the mother of each rule as the category for which we are trying to find a FIRST value. Throughout each iteration, unification of a daughter with the lhs of an element of FIRST results in a modified rule and a modified pair in which bindings between the mother category and the rhs of the pair are established. The modified mother and rhs are then used to construct the pair which is added to FIRST. For instance, given rule $X \rightarrow Y$ and pair $(L, R)$, we unify $Y$ and $L$ to give $X' \rightarrow Y'$ and $(L', R')$; from these the pair $(X', R')$ is constructed and added to FIRST.

The algorithm assumes an operation $+_\leq$ which constructs a set $S' = S +_\leq p$ in the following way: if pair $p$ subsumes an element $a$ of $S$ then $S' = S - a + p$; if $p$ is subsumed by an element of $S$ then $S' = S$; else $S' = S + p$. It should be noted that the pairs constituting the value of FIRST can themselves be compared using the subsumption relation in which reentrant values are subsumed by non-reentrant ones, and combined using the unification operation. Thus in the principal step of the algorithm, a new pair is constructed as described above, a restrictor is applied to it, and the resulting, restricted pair is $+_\leq$-added to FIRST. The algorithm is as follows:

1. Initialise $First = \{\}$.

2. Run through all the daughters in the grammar. If $X$ is pre-terminal, then $First = First +_\leq (X,X)!\Phi$ (where $(X,X)!\Phi$ means apply the negative restrictor $\Phi$ to the pair $(X,X)$).

3. For each rule in the grammar with mother

| S | $\Rightarrow$ **NP**[agr: X, slash: NULL] **VP**[agr: X, slash: NULL] |
| --- | --- |
| S | $\Rightarrow$ **NP**[slash: NULL] **NP**[agr: X, slash: NULL] **VP**[agr: X, slash: NP] |
| **VP**[agr: X, slash: Y] | $\Rightarrow$ **Vtra**[agr: X, ter: +] **NP**[slash: Y] |
| **NP**[agr: X, slash: NULL] | $\Rightarrow$ **Det**[ter: +] **N**[agr: X, ter: +] |
| **NP**[slash: NP] | $\Rightarrow \epsilon$ |

Figure 1: Example grammar with value sharing.

$X$, apply steps 4 and 5 until no more changes are made to $First$.

4. If the rule is $X \rightarrow \epsilon$, then $First = First +_{\leq} (X, \epsilon)!\Phi$.

5. If the rule is $X \rightarrow Y_1..Y_i..Y_k$, then $First = First +_{\leq} (X', a)!\Phi$ if $(Y_i', a)$ has successfully unified with an element of $First$, and $(Y_1', \epsilon_1)...(Y_{i-1}', \epsilon_{i-1})$ have all successfully and simultaneously unified with members of $First$. Also, $First = First +_{\leq} (X', \epsilon)!\Phi$ if $(Y_1', \epsilon_1)...(Y_k', \epsilon_k)$ have all successfully and simultaneously unified with elements of $First$.

6. Now, for any string of categories $X_1..X_i..X_n$, $First = First +_{\leq} (X_1'...X_n', a)!\Phi$ if $(X_1', a)$ has successfully unified with an element of $First$, and $a \not\leq \epsilon$. Also, for $i = 2...n$, $First = First +_{\leq} (X_1'...X_n', a)!\Phi$ if $(X_i', a)$ has successfully unified with an element of $First$, $a \not\leq \epsilon$, and $(X_1', \epsilon_1)...(X_{i-1}', \epsilon_{i-1})$ have all successfully and simultaneously unified with members of $First$. Finally, $First = First +_{\leq} (X_1'...X_n', \epsilon)!\Phi$ if $(X_1', \epsilon_1)...(X_n', \epsilon_n)$ have all successfully and simultaneously unified with members of $First$. (This step may be computed on demand).

One observation on this algorithm is in order. The last action of steps 5 and 6 adds $\epsilon$ as a possible value of FIRST for a mother category or a string of categories; such a value results when all daughters or categories have $\epsilon$ as their FIRST value. Since most grammatical descriptions assign a category to $\epsilon$ (e.g. to bind onto it information necessary for correct gap threading), the pairs $(X', \epsilon)$ or $(X_1'...X_n', \epsilon)$ should have bindings between their two elements; this creates the problem of deciding which of the $\epsilon$s in the FIRST pairs to use, since it is possible in principle that each of these will have a different value for $\epsilon$. In our implementation, the pair added to $First$ in these situations consists of the mother category or the string of categories and the most general category for $\epsilon$ as defined by the grammar, thus effectively ignoring any bindings that $\epsilon$ may have within the constructed pair. A more accurate solution would have been to compute multiple pairs with $\epsilon$, construct their least upper bound, and then add this to $First$. However, in our implementation this solution has not proven necessary.

### 4.2 EXAMPLE

Assuming the grammar in Fig. 1 and the negative restrictor $\Phi = \{\text{slash}\}$, the following is a simplified run through the algorithm:

• $First = \{\}$

• After processing all pre-terminal categories $First = \{(Det, Det), (N, N), (Vtra, Vtra)\}$ (obvious bindings not shown).

• After the first iteration $First = \{(Det, Det), (N, N), (Vtra, Vtra), (VP[agr : X], Vtra[agr : X]), (NP, Det), (NP, \epsilon)\}$

• Since 'slash' is in $\Phi$, any of the NPs in the grammar will unify with the lhs of $(NP, \epsilon)$ and hence S will have Vtra as part of its FIRST value. $First = \{.., (VP[agr : X], Vtra[agr : X]), (NP, Det), (NP, \epsilon), (S, Det), (S, Vtra)\}$

• The next iteration adds nothing and the first stage of the algorithm terminates.

The second stage (step 6) is done on demand, for example to compute state transitions for a parsing table, in order to avoid the expense of computing FIRST for all possible substrings of categories. For instance, to compute FIRST for the string [NP NP VP] the algorithm works as follows:

- $First = \{..,(VP[agr : X], Vtra[agr : X]),$
$(NP, Det), (NP, \epsilon)...\}$

- After considering the first NP: $First = \{..,([NP\ NP\ VP], Det)\}$.

- Consideration of the second NP in the input string results in no changes to $First$, given the semantics of $+_\leq$, since the pair that it would have added, $([NP\ NP\ VP], \epsilon)$, is already in $First$.

- Since NPs can rewrite as $\epsilon$ (i.e. $(NP, \epsilon)$ is in $First$), $First = \{..,([NP\ NP\ VP], Det),$
$([NP\ NP\ VP], Vtra)\}$.

- Finally, $([NP\ NP\ VP], \epsilon)$ may not be added since $(VP, \epsilon)$ does not unify with any element of $First$.

## 5 Improving the Search Through $First$

If the algorithm is run as presented, each iteration through the grammar rules becomes slower and slower. The reason is that, in step 5, when searching $First$ to create a new pair $(X', a)$, every pair in $First$ is considered and unification of its lhs with the relevant daughter of $X$ attempted. Since each iteration normally adds pairs to $First$ each iteration involves a search through a larger and larger set; furthermore, this search involves unification, and in the case of a successful match, the subsequent construction and addition to $First$ also requires subsumption checks. All of these operations combine to make each additional element in $First$ have a strong effect on the performance of the algorithm. We therefore need to minimize the number of pairs searched.

Considering the dependencies that exist between pairs in $First$ one notices that once a pair has been considered in relation with all the rules in the grammar, the effect of that pair has been completely determined. That is, after a pair is added to $First$ it need only be considered up to and including the rule from which it was derived, after which time it may be excluded from further searches. For example, take the previous grammar, and in particular the value of $First$ after the first iteration through the algorithm. The pair $(NP, Det)$, added because of the rule **NP**[agr: X, slash: NULL] $\Rightarrow$ **Det**[ter: +] **N**[agr: X, ter: +], has to be considered only once by every rule in the grammar; after that, this pair cannot be involved in the construction of new values.

A simple data structure which keeps track of those pairs that need to be searched at any one time was added to the algorithm; the data structure took the form of a list of pointers to active pairs in $First$, where an active pair is one which has not been considered by the rule from which it was constructed. For example, the pair $(NP, Det)$ would be active for a complete iteration from the moment that the corresponding rule introduced it until that rule is visited again during the second iteration. The effect of this policy is to allow each pair in $First$ to be tested against each rule exactly once and then be excluded from subsequent searches; this greatly reduces the number of pairs considered for each iteration.

Using the Typed Feature Structure system (the LKB) of Briscoe et al. (1993), we wrote two grammars and tested the algorithm on them. Table 1 shows the average number of pairs considered for each iteration compared to the average number of pairs in $First$.

|         | 13 Rule Grammar | | 21 Rule Grammar | |
|---------|------------|-------|------------|-------|
|         | Considered | Total | Considered | Total |
| Iter. 1 | 3.5        | 3.5   | 8.4        | 8.4   |
| Iter. 2 | 7.5        | 10.7  | 9.7        | 18.7  |
| Iter. 3 | 1.2        | 12.0  | 1.0        | 19.0  |

Table 1: Average number of pairs per iteration.

As we can see, after the first iteration the number of pairs that needs to be considered is less (much less for the final iteration) than the total number of pairs in $First$. Similar improvements in performance were obtained for the computation of FOLLOW.

## 6 Related Research

The extension to the LR algorithm presented by Nakazawa (1991) uses a similar approach to that described here; the functions involved however are those necessary for the construction of an LR parsing table (i.e. the GOTO and ACTION functions). One technical dif-ference between the two approaches is that he uses positive restrictors (Shieber 1985) instead of negative ones. In addition, both of his algorithms also differ in another way from the algorithm described here. The difference is that they add items to a set using simple set addition whereas in the algorithm of Section 4.1 we add elements using the operator $+_{\leq}$. Furthermore, when computing the closure of a set of items, both of the algorithms there ignore the effect that unification has on the categories in the rules.

For example, the states of an LR parser are computed using the closure operation on a set $I$ of dotted rules or items. In Nakazawa's algorithms computation of this closure proceeds as follows: if dotted rule $< A \rightarrow w.Bx >$ is in $I$, then add a dotted rule $< C \rightarrow .y >$ to the closure of $I$, where $C$ and $B$ unify. This ignores the fact that both dotted rules may be modified after unification, and therefore, his algorithm leads to less restricted $I$ values than those implicit in the grammar. To adapt our algorithm to the computation of the closure of $I$ for a feature-theoretic grammar would involve using a set of pairs of dotted rules as the value of $I$.

# 7 CONCLUSION

We have extended an algorithm that manipulates CF grammars to allow it to handle feature-theoretic ones. It was shown how most of the information contained in the grammar rules may be preserved by using a set of pairs as the value of a function and by using the notion of subsumption to update this set. Although the algorithm has in fact been used to adapt the constraint propagation algorithm of Brew (1992) to phrase structure grammars, the basic idea should be applicable to the rest of the functions needed for constructing LR tables. However, such adaptations are left as a topic for future research.

Finally, improvements in speed obtained with the active pairs mechanism of Section 5 are of an order of magnitude in an implementation using Common Lisp.

# ACKNOWLEDGEMENTS

This work was funded by the UK SERC. I am very grateful to Ted Briscoe, John Carroll, Mark-Jan Nederhof, Ann Copestake and two anonymous reviewers. All remaining errors are mine.